\documentclass[11pt,a4paper]{article}
\usepackage[hyperref]{informative_captions}
\usepackage{times}
\usepackage{latexsym}

\usepackage{multirow}
\usepackage{amsmath}
\usepackage{graphicx}
\usepackage{booktabs}
\usepackage{flexisym}
\usepackage{enumitem}
\usepackage{float}
\usepackage{array}

\usepackage{url}

\aclfinalcopy 
\def\aclpaperid{1864} 

\newcommand{\LabelType}[1]{\textsc{\textcolor{blue}{$\langle$#1$\rangle$}}}

\newcolumntype{C}[1]{>{\centering\arraybackslash}p{#1}}

\def\CovE{Cov^r_{we}}
\def\CovO{Cov^p_{obj}}
\def\WeE{W^r_{we}}
\def\WeO{W^p_{obj}}

\title{Informative Image Captioning with External Sources of Information}

\author{\hspace*{-40mm}Sanqiang Zhao \\ \hspace*{-40mm}University of Pittsburgh \\ \hspace*{-40mm}{\tt \small{sanqiang.zhao@pitt.edu}} \\\And
        \hspace*{-20mm}Piyush Sharma~~~~~~~Tomer Levinboim~~~~~~~Radu Soricut \\ \hspace*{-20mm}Google Research, Venice, CA 90291 \\ \hspace*{-20mm}{\tt \small{\{piyushsharma,tomerl,rsoricut\}@google.com}}}

\date{}

\begin{document}
\setlength{\abovedisplayskip}{3pt}
\setlength{\belowdisplayskip}{3pt}
\setlength{\abovedisplayshortskip}{3pt}
\setlength{\belowdisplayshortskip}{3pt}
\setlength{\belowcaptionskip}{3pt}
\setlength{\abovecaptionskip}{3pt}
\maketitle

\begin{abstract}
An image caption should fluently present the essential information in a given image, including informative, fine-grained entity mentions and the manner in which these entities interact.
However, current captioning models are usually trained to generate captions that only contain common object names, thus falling short on an important ``informativeness'' dimension.
We present a mechanism for integrating image information together with fine-grained labels (assumed to be generated by some upstream models) into a caption that describes the image in a fluent and informative manner.
We introduce a multimodal, multi-encoder model based on Transformer that ingests both image features and multiple sources of entity labels.
We demonstrate that we can learn to control the appearance of these entity labels in the output, resulting in captions that are both fluent and informative.
\end{abstract}

\section{Introduction}
\label{sec:intro}

\begin{figure}[t!]
  \centering
  \scriptsize
  \renewcommand{\arraystretch}{1.5}
  \begin{tabular}{l l l}
    \raisebox{-.5\height}{\includegraphics[width=0.35\linewidth]{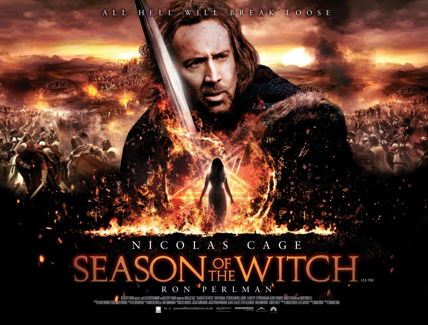}} &
    \begin{tabular}{p{0.5\linewidth}}
        \textbf{Baseline Model:} ``return to the main poster page for film'' \\
        \textbf{Our Approach:} ``extra large movie poster image for \textcolor{olive}{Season of the Witch}''\\
    \end{tabular} \\

    \raisebox{-.5\height}{\includegraphics[width=0.35\linewidth]{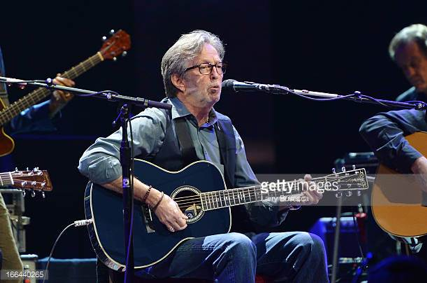}} &
    \begin{tabular}{p{0.5\linewidth}}
        \textbf{Baseline Model:} ``folk rock artist performs on stage during festival'' \\
        \textbf{Our Approach:} ``\textcolor{olive}{Eric Clapton} performs on stage during the \textcolor{olive}{Crossroads Guitar Festival}''\\
    \end{tabular} \\


  \end{tabular}
  \caption{Generating informative captions using fine-grained entity information from external sources; baseline outputs from \newcite{sharma2018conceptual}.}
  \label{fig:sample}
\end{figure}

Much of the visual information available on the web is in the form of billions of images, but that information is not readily accessible to those with visual impairments, or those with slow internet speeds.
Automatic image captioning can help alleviate this problem, but its usefulness is directly proportional to how much information an automatically-produced caption can convey.
As it happens, the goal of learning good models for image captioning (in terms of generalization power) is at odds with the goal of producing highly informative captions (in terms of fine-grained entity mentions).
For this reason, previous approaches to learning image captioning models at web scale~\cite{sharma2018conceptual} had to compromise on the informativeness aspect, and trained models that could not produce fine-grained entity mentions (e.g., ``Season of the Witch'') and instead settled for the conceptual (i.e., hypernym) variant for such entities (e.g., ``film'') as shown in Fig.~\ref{fig:sample}.
We present an approach that solves this problem by leveraging upstream models that are capable of producing fine-grained entity names, and integrating them in a controlled manner to produce captions that are both fluent and highly informative.

The standard approach to the image captioning task uses $\langle image, caption\rangle$ pairs to train an image-to-text encoder-decoder model.
The ``image encoder'' is usually a Convolutional Neural Network that extracts image features.
The ``text decoder'' is usually a Recurrent Neural Network or a Transformer Network~\cite{vaswani2017attention} that depends solely on these image features to generate the target caption.
We identify two limitations of this approach that restrict the amount of information that the generated captions contain:
\begin{enumerate}[noitemsep]
  \item Fine-grained entity recognition is a challenging task in itself, and solving it requires specialized datasets and models.
        Attempts to simultaneously recognize fine-grained entities and generate an informative caption have previously failed (for example, see~\newcite{sharma2018conceptual}, Fig.~3 therein).
        In addition, image metadata may be available and requires models capable of smoothly incorporating it.
  \item The $\langle{image, caption}\rangle$ pairs on which such models are trained usually have $caption$ capturing only a limited coverage of the entities present in the image (or its meta-information).
        At training time, this limitation gets baked into the models and inherently limits the amount of information presented in the output caption at inference time.
\end{enumerate}

To address the above shortcomings, we define the caption generation task as a new $\langle image, entities\rangle$-to-caption task focused on fluently incorporating entities in the generated caption.
We opt for an approach in which entity labels produced by some upstream model(s) are consumed as inputs to the captioning model, in addition to the image pixels.
This allows us to use off-the-shelf image labeler models (for object labels, entity recognition, etc.), trained specifically for accuracy on their tasks.
To address the second limitation above, we introduce a modeling mechanism that allows us to learn (at training time) and control (at inference time) the coverage of entity mentions in the generated captions.
From a modeling perspective, we contribute along these lines by introducing
\begin{enumerate}[noitemsep]
  \item a multi-encoder model architecture that, paired with a multi-gated decoder, integrates image-based information with fine-grained entity information and allows us to generate entity-rich captions
  \item a coverage control mechanism that enables us to learn how to control the appearance of fine-grained entities in the generated caption at inference time.
\end{enumerate}
Furthermore, we perform empirical evaluations using both automatic metrics and human judgments, and show that the approach we propose achieves the effect of boosting the informativeness and correctness of the output captions without compromising their fluency.

\section{Related Work}
Automatic image captioning has a long history, starting with earlier work~\citep{hodosh13framing,donahue2014long,karpathy2014deep,kiros2014unifying}, and continuing with models inspired by sequence-to-sequence models~\citep{sutskever2014sequence,bahdanau2014neural} adapted to work using CNN-based image representations (\citep{vinyals2015show,fang2015captions,xu2015show,mixer15,yang2016review,liu2017optimization}, etc.).
As training data, the MS-COCO~\cite{coco} is the most used dataset, while the Conceptual Captions dataset \cite{sharma2018conceptual} is a more recent web-centric, large volume resource.

The work in \newcite{you2016image, yao2017boosting} is related to our approach, as they integrate precomputed attributes into image captioning models.
These attributes guide the model to generate captions with correct objects, and are obtained from upstream object detection models that use fairly coarse-grained object labels.
Even closer, \newcite{lu2018entity} propose an approach for incorporating fine-grained entities by generating a ``template'' caption with fillable slots.
They replace entity names in the data with a slot that indicates which entity type should be used to fill that slot, and use a postprocessing step to replace the type slot with the entity name.

The work we present here is novel both with respect to the data preparation and the proposed model.
For data, we operate at web-scale level by enhancing Conceptual Captions \cite{sharma2018conceptual} (3.3M images) with fine-grained annotations.
For modeling, we describe a framework that extends Transformer Networks~\cite{vaswani2017attention}, and allows for the principled integration of multiple, multimodal input signals.
This framework allows us to test a variety of experimental conditions for training captioning models using fine-grained labels.
In addition, our framework has a coverage control mechanism over fine-grained label inputs, which can differentiate between labels for which we need high-recall and labels for which we desire high-precision.

\section{Data and Models}
\subsection{Data Preparation}
\label{sec:data_prep}
The goal of this stage is to obtain annotations that contain (i) entity-rich captions as ground-truth and, (ii) entities associated with each image using fine-grain label detectors.
To that end, we build on top of the Conceptual Captions dataset \cite{sharma2018conceptual}, containing 3.3~Million $\langle image, caption \rangle$ pairs.
For Conceptual Captions, the ground-truth $caption$ is obtained by substituting fine-grained entity mentions in Alt-text\footnote{https://en.wikipedia.org/wiki/Alt\_attribute} with their corresponding hypernyms (e.g., ``Los Angeles'' is substituted by ``city'').
Although this simplification makes the captions more amenable to learning, it leads to severe loss of information.
To achieve goal (i) above, we reprocessed the URLs from Conceptual Captions and remapped the hypernyms back to their corresponding fine-grained entities (e.g., map ``city'' back to ``Los Angeles''), using the surrounding text as anchors.

\subsubsection{Fine-grained Image Labels}
To achieve goal (ii) above, we employ pretrained models to extract from input images (1) object detection labels and, (2) web entity labels, using Google Cloud Vision APIs.\footnote{https://cloud.google.com/vision Google Cloud Vision API uses Google Image Search to find topical entities like celebrities, logos, or news events.}
Object labels refer to fine-grained common objects (e.g., ``eucalyptus tree'' and ``sedan'').
Web entity labels, on the other hand, refer to fine-grained named entities (e.g., ``Los Angeles'' and ``Toyota'').
In addition to the image pixels, these labels serve as inputs, and during training the model needs to learn a mapping between these labels and the corresponding fine-grained entities in the ground-truth captions.

\subsubsection{Selective Hypernym Substitution}
An additional issue that needs to be resolved in this data preparation stage is that there is no guarantee of a complete mapping between the fine-grained labels and web entities in the input and the ones in the output (the former are produced by models, while the latter are coming from human-authored Alt-text).
Training a model in which output fine-grained entities are not present in the additional input labels is problematic, because it would again require the model to perform both fine-grained entity recognition from pixels as well as caption generation (known to result in hallucination and mis-identification issues, see \newcite{sharma2018conceptual}).

To avoid this pitfall, we apply ``selective hypernymization'' (in the same vein as~\newcite{sharma2018conceptual}), for which we retain a fine-grained entity in the ground-truth caption only if it is present in the input labels; otherwise, we substitute it by its corresponding hypernym (if present in the ground-truth) or remove it entirely (if not).
This step ensures that the data contains a surjective mapping for the fine-grained labels between input and output labels, resulting in learnable mappings between input and output fine-grained labels.
For example, in Fig.~\ref{fig:sample}, the raw Alt-text is \emph{``Eric Clapton performs on stage during the 2013 Crossroads Guitar Festival at Madison Square Garden''}.
The additional input label are \emph{``Eric Clapton''}, \emph{``Musician''} and \emph{``Crossroads Guitar Festival 2013''}.
To ensure the surjective property of the fine-grained label mapping, the mention \emph{``Madison Square Garden''} is removed, resulting in the ground-truth \emph{``Eric Clapton performs on stage during the 2013 Crossroads Guitar Festival''}.
Note that we do not enforce a fully bijective mapping between the labels, and may have input labels with no correspondence in the output; for these instances, the model needs to learn that they should not be covered.


\subsection{Models}

\subsubsection{Multi-Encoder Transformer}
\label{sec:multi-enc-transformer}

\begin{figure}[t]
\begin{center}
 \includegraphics[width=1.0\linewidth]{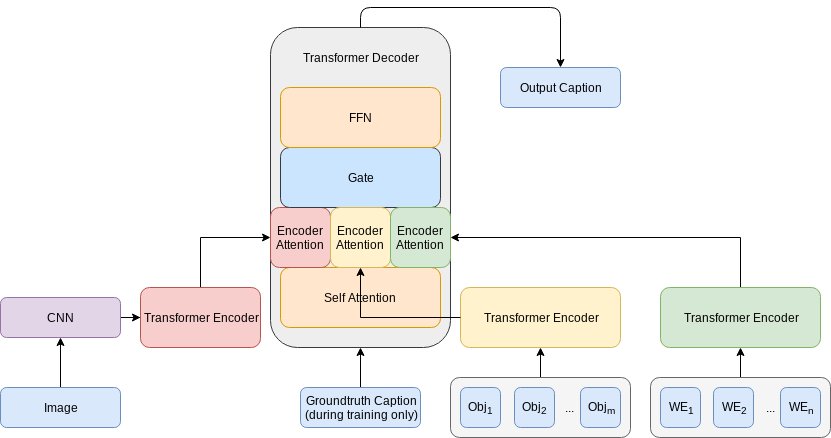}
\end{center}
 \caption{A multi-encoder Transformer Network processes the input image, object labels and web entity labels. The shared decoder attends to all encoders' outputs and combines their information.}
\label{fig:multi_enc_t2t}
\end{figure}

We introduce a multi-encoder extension to Transformer Networks \cite{vaswani2017attention} that is used to process our multimodal inputs: image features, object labels, and web entity labels (Fig.~\ref{fig:multi_enc_t2t}).
Self-attention layers in the Transformer encoder help with learning label representations in the context of the other labels.

\paragraph{Image Encoder}
To encode the image information, we use a Convolutional Neural Network (CNN) architecture to extract dense image features ($\textbf{Img}=\{img_1,img_2,...,img_k\}$) corresponding to a uniform grid of image regions.
A Transformer-based encoder takes these image features and embeds them into the features space shared by all input modalities, $\textbf{H}_{img}$ = $f_{enc}(\textbf{Img}$, $\theta_{enc\_img}$), where $\theta_{enc\_img}$ refers to the parameters for this image encoder.

\begin{figure}[h]
\begin{center}
 \includegraphics[width=0.7\linewidth]{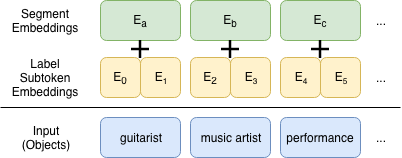}
\end{center}
 \caption{Learnable representations for the Object labels using their surface tokens.}
\label{fig:obj_enc}
\end{figure}

\paragraph{Object Label Encoder}
The input for this encoder is an ordered sequence of object labels, sorted by the confidence score of the model that predicts these labels.
This allows the model to learn that labels appearing at the head of the sequence are more reliable.
For each separate label, we create learnable segment embeddings (inspired by \newcite{devlin2018bert}) as shown in Fig.~\ref{fig:obj_enc}, using the subtokenization scheme described in \newcite{sennrich2015neural}.
A Transformer-based encoder network takes these object label features, $\textbf{Obj}=\{obj_1,obj_2,...,obj_m\}$, and embeds them into the features space shared by all input modalities, $\textbf{H}_{obj} = f_{enc}(\textbf{Obj}, \theta_{enc\_obj}$), where $\theta_{enc\_obj}$ refers to the parameters for this object encoder.
We do not apply positional embeddings because the relative positions of object labels are irrelevant.

\paragraph{Web Entity Label Encoder}
For modeling web entity labels, we experiment with two modeling variants that consider either (i) the web entity type, or (ii) the web entity surface tokens.
For (i), we obtain entity types by using the Google Knowledge Graph (KG) Search API to match the web entity names to KG entries.
Each of these types is subsequently represented by a trainable embedding vector.
The model is trained to predict captions with entity types, which during post-processing are substituted by the highest scored web entity label of the predicted type.
If no such typed label exists, we use the generic name of the type itself (e.g., ``film'').

\begin{figure}[h]
\begin{center}
 \includegraphics[width=0.7\linewidth]{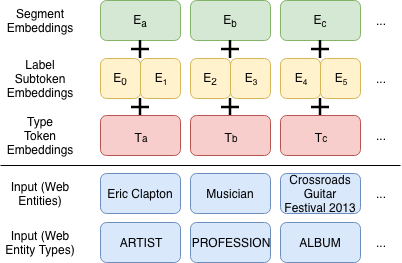}
\end{center}
 \caption{Learnable representations for the Web Entity labels using the surface tokens (and their types).}
\label{fig:we_nontemplate}
\end{figure}

For variant (ii), the model directly attempts to model and generate a caption containing the surface tokens.
In this case, the entity type is still provided as an input to the model as additional source of information.
These input representations are constructed by summing up a trainable segment embedding with the subtoken embeddings and the type embedding (Fig.~\ref{fig:we_nontemplate}).
A Transformer encoder network takes these web entity features, $\textbf{WE}=\{we_1,we_2,...,we_n\}$, and
embeds them into the feature space shared by all input modalities, $\textbf{H}_{we} = f_{enc}(\textbf{WE}, \theta_{enc\_we}$), where $\theta_{enc\_we}$ refers to the parameters for the web entity encoder.
Similar to object labels, positional embeddings are not applied to the sequence of web entity labels.

\subsubsection{Multi-gated Transformer Decoder}
To accommodate the multi-encoder architecture on the input side, we propose a multi-gated extension to the Transformer decoder~\cite{vaswani2017attention}.
As usual, this decoder is a stack of $k$ identical layers, where each of these layers has 3 sub-layers: a self-attention layer, an encoder-attention layer, and a fully connected feed-forward layer.
Among the 3 sub-layers, we modify the encoder-attention sub-layer by introducing a mechanism that combines information coming from different encoders.

Formally, we denote the hidden states of $n$-th layer by $\textbf{Z}_n = {z_{n,1}, ... , z_{n,T} }$ ($\textbf{Z}_0$ refers to decoder input embeddings, and $T$ is the length of decoder inputs).
The self-attention sub-layer equation is given in Eq.~\ref{eq:decoder_selfattn}; as expected, the inputs to this layer are masked to the right, in order to prevent the decoder from attending to ``future'' positions (i.e., $z_{n,j}$ does not attend to $z_{n,j+1}, ... , z_{n,T}$).
\begin{align}
\fontsize{8}{10}\selectfont
\textbf{Z\textprime}_{n,j} = \text{SelfAttn}(z_{n,j}, \textbf{Z}_{n,1:j}, \theta_{self\_attn})
\label{eq:decoder_selfattn}
\end{align}

Next, the encoder-attention sub-layer contains three attention modules, which enables it to attend to the three encoder outputs (Eq.~\ref{eq:decoder_encattn}):
\begin{align}
\fontsize{8}{10}\selectfont
\begin{split}
\textbf{Z\textprime\textprime}_{n, j}^{img} & = \text{EncAttn}(z\textprime_{n,j}, \textbf{H}_{img}, \theta_{enc\_attn\_img})
\\
\textbf{Z\textprime\textprime}_{n, j}^{obj} &= \text{EncAttn}(z\textprime_{n,j}, \textbf{H}_{obj}, \theta_{enc\_attn\_obj})
\\
\textbf{Z\textprime\textprime}_{n, j}^{we} &= \text{EncAttn}(z\textprime_{n,j}, \textbf{H}_{we}, \theta_{enc\_attn\_we})
\end{split}
\label{eq:decoder_encattn}
\end{align}

We expect the model to have the ability to adaptively weight each of these three source of information, and therefore we introduce a multi-gate sub-layer.
For each source $S$, we compute a gate $Gate_{n,j}^S$ value that determines the amount of information that flows through it (Eq.~\ref{eq:decoder_gate}).
Each gate value is computed by transforming the concatenate of the outputs from the three encoder attentions.

\begin{align}
\fontsize{8}{10}\selectfont
\label{eq:decoder_gate}
\begin{split}
Gate_{n,j}^{img} &= \text{tanh}(\textbf{U}_{img} * \text{concat}(\textbf{Z\textprime\textprime}_{n, j}^{img};\textbf{Z\textprime\textprime}_{n, j}^{obj};\textbf{Z\textprime\textprime}_{n, j}^{we}))
\\
Gate_{n,j}^{obj} &= \text{tanh}(\textbf{U}_{obj} * \text{concat}(\textbf{Z\textprime\textprime}_{n, j}^{img};\textbf{Z\textprime\textprime}_{n, j}^{obj};\textbf{Z\textprime\textprime}_{n, j}^{we}))
\\
Gate_{n,j}^{we} &= \text{tanh}(\textbf{U}_{we} * \text{concat}(\textbf{Z\textprime\textprime}_{n, j}^{img};\textbf{Z\textprime\textprime}_{n, j}^{obj};\textbf{Z\textprime\textprime}_{n, j}^{we}))
\end{split}
\end{align}
\noindent
The output of gate sub-layer is a soft switch that controls the information flow from the three encoders (Eq.~\ref{eq:decoder_gate_com}):
\begin{align}
\fontsize{8}{10}\selectfont
\label{eq:decoder_gate_com}
\begin{split}
\textbf{Z\textprime\textprime\textprime}_{n, j} = &~Gate_{n,j}^{img} * \textbf{Z\textprime\textprime}_{n, j}^{img} \\
 &+ ~Gate_{n,j}^{obj} * \textbf{Z\textprime\textprime}_{n, j}^{obj} \\
 &+ ~Gate_{n,j}^{we} * \textbf{Z\textprime\textprime}_{n, j}^{we}
\end{split}
\end{align}
\noindent
Finally, as in the vanilla Transformer decoder, the third sub-layer is a feed-forward network that processes the representation for the next $n$+1 layer:
\begin{align}
\fontsize{8}{10}\selectfont
\label{eq:decoder_ffn}
\textbf{Z}_{n+1, j} = \text{FFN}(\textbf{Z\textprime\textprime\textprime}_{n, j})
\end{align}

The three sources of information (image, object labels, and web entity labels) are treated symmetrically in the above equations.
However, the only ``true'' source of information in this case is the image, whereas the other labels are automatically-produced annotations that can vary both in quality and other properties (e.g., redundancy).
We introduce an asymmetry in the modeling that will allow us to capture this important distinction.

\subsubsection{Label Coverage Control}

Because of the asymmetry between the image input and the label inputs, we introduce a mechanism to control the coverage of the supplied object and web entity labels in the generated caption.
This mechanism consists of two parts: (i) two regressor models that learn coverage scores correlating input labels (one for objects, one for web entities) with output mentions, and (ii) two control ``knobs'' that allow us to specify desired coverage scores at inference time.
This coverage control mechanism is inspired by the Label-Fine-Tuning model of~\newcite{niu2018polite}, although it is used here to achieve a different goal.

\paragraph{\emph{Coverage of Object Labels}}
An interesting property of the object labels is that they may be repetitive, often at various levels of granularity, for instance ``table'', ``office table'' and ``office''.
A model that would require to reproduce all of them in the output caption will likely produce a disfluent caption containing repetitive mentions of the same object.
We introduce object level coverage as a precision-like score for object labels, $\CovO$, defined as the fraction of output caption tokens present in the input object labels (Eq.~\ref{eq:coverage}).

\paragraph{\emph{Coverage of Web Entities}}
In contrast with the object labels, web entity labels are not repetitive, and tend to have high information value.
For that reason, we want a high fraction of input web entities to be used in the output caption.
Therefore, we introduce the web entity coverage as a recall-like score for web entity labels, $\CovE$, defined as the fraction of input tokens that are present in the caption (Eq.~\ref{eq:coverage}).
\begin{align}
\fontsize{8}{10}\selectfont
\label{eq:coverage}
\begin{split}
\CovO &= \frac{|\{\text{objects~tokens}\} \cap \{\text{caption~tokens}\}|}{|\{\text{caption~tokens}\}|} \\
\CovE &= \frac{|\{\text{entities~tokens}\} \cap \{\text{caption~tokens}\}|}{|\{\text{entities~tokens}\}|}
\end{split}
\end{align}

\begin{figure}[t]
\begin{center}
  \includegraphics[width=1.0\linewidth]{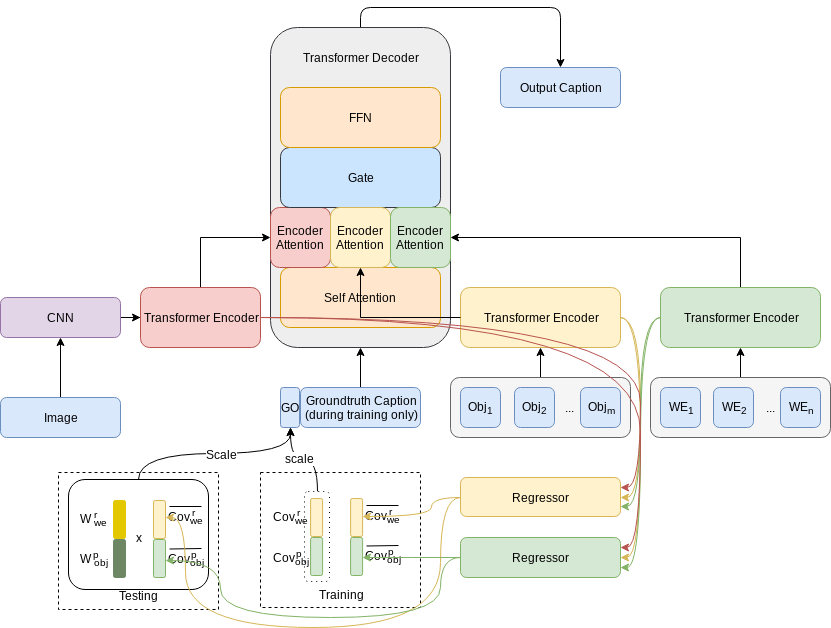}
\end{center}
  \caption{The Image Transformer Encoder (left side) is the only ``true'' source of information; the Transformer Encoders on the right side encode model-produced sources of information.
  The control mechanism (two Regressor models) learns this asymmetry (during training), and provides additional run-time control. }
\label{fig:coverage_controller_comb}
\end{figure}

\subsubsection{Label Coverage Prediction \& Control}
\label{sec:cov_pred}
We train two regressors to predict the coverage scores for object labels ($\overline{\CovO}$) and web entity labels ($\overline{\CovE}$), using as features the outputs of the Transformer encoders (Eq.~\ref{eq:pred}).
At training time, ground-truth captions are known, so the regression target values $\CovO$ and $\CovE$ are computed using Eq.~\ref{eq:coverage}.
When training regressors parameters ($\textbf{U}^p_{obj}$ and $\textbf{U}^r_{we}$), we fix the Transformer parameters and minimize the regression losses (Eq.~\ref{eq:reg}).
\begin{align}
\fontsize{8}{10}\selectfont
\label{eq:pred}
\begin{split}
\overline{\CovO} &= \text{sigmoid}(\textbf{U}^p_{obj} \text{concat}(\textbf{H}_{img};\textbf{H}_{obj};\textbf{H}_{we})) \\
\overline{\CovE} &= \text{sigmoid}(\textbf{U}^r_{we} \text{concat}(\textbf{H}_{img};\textbf{H}_{obj};\textbf{H}_{we}))
\end{split}
\end{align}

\begin{align}
\fontsize{8}{10}\selectfont
\label{eq:reg}
\begin{split}
loss_{obj}^{reg} &= (\CovO - \overline{\CovO})^2 \\
loss_{we}^{reg} &= (\CovE - \overline{\CovE})^2
\end{split}
\end{align}

We compose a coverage indicator vector of the same dimensionality as the word embeddings by tiling the two coverage scores, and use this coverage indicator vector to scale (element-wise) the start token of the output caption.
At training time, we use the actual coverage scores, $\CovE$ and $\CovO$ (see `Training' box, the lower part of Fig.~\ref{fig:coverage_controller_comb}).
At run-time, we use the scores predicted by the regressors, $\overline{\CovE}$ and $\overline{\CovO}$ (box labeled `Testing' in Fig.~\ref{fig:coverage_controller_comb}), which we can additionally scale using the two scalars $\WeE$ and $\WeO$.
These additional scalars act as coverage boost factors and allows us to control, at inference time, the degree to which we seek increased coverage and therefore obtain captions that are both fluent and more informative (by controlling $\WeE$ and $\WeO$).

\section{Experiments}
\paragraph{Dataset}
We extended the Conceptual Captions dataset as described in Section~\ref{sec:data_prep}.
We use the standard (v1.0) splits with 3.3M training samples, and approximately 28K each for validation and test.
The human evaluations use a random sample of 2K images from the test set.

\paragraph{Image Processing}
In this work, we use ResNet~\cite{he2016deep} for processing the image pixels into features (output features size 7x7x2048), pretrained on the JFT dataset \cite{hinton2015distilling}.\footnote{This configuration performed best among the CNN and pretraining conditions we evaluated against.}
Input images undergo random perturbations and cropping before the CNN stage for better generalization.

\paragraph{Text Handling}
We use subtoken embeddings \cite{sennrich2015neural} with a maximum vocabulary size of 8k for modeling caption tokens, web entities and object labels.
Captions are truncated to 128 tokens.
We use an embedding size of 512, with shared input and output embeddings.

\paragraph{Model Specification}
We use 1 layer and 4 attention heads for Web Entity label encoder; 3 layers and 1 attention head for Object label encoder; 1 layer and 4 attention heads for the CNN encoder; and 6 layers and 8 heads for the shared decoder.

\paragraph{Optimization}
MLE loss is minimized using Adagrad \cite{duchi2011adaptive} with learning rate 0.01 and mini-batch size 32.
Gradients are clipped to global norm 4.0.
We use 0.2 dropout rate on image features to avoid overfitting.
For each configuration, the best model is selected to maximize the CIDEr score on the development set.

\paragraph{Inference}
During inference, the decoder prediction of the previous position is fed to the input of the next position.
We use a beam search of size 4 to compute the most likely output sequence.

\subsection{Quantitative Results}


We measure the performance of our approach using three automatic metrics (see Eq.~\ref{eq:coverage}):
\begin{description}
\item[CIDEr] measures similarity between output and ground-truth \cite{vedantam2015cider}.
\item[Web-Entity coverage] {\bf$\CovE$} measures the recall of input web entity labels in the generated caption.
\item[Object coverage] {\bf$\CovO$} measures the precision of the output caption tokens w.r.t. input object labels.
\end{description}
To measure how well our model combines information from \emph{both} modalities of inputs (i.e. image and entity labels), we compare its peformance against several baselines:

\begin{description}
\item[Image only]: \newcite{anderson2018bottom} (using Faster R-CNN trained on Visual Genome) and \newcite{sharma2018conceptual} (using ResNet pretrained on JFT)
\item[Entity-labels only]:
  Transformer model trained to predict captions from a sequence of entity labels
  (vanilla Transformer encoder/decoder with 6 layers and 8 attention heads).
\item[Image\&Entity-labels]: \newcite{lu2018entity} (w/ Transformer), their template approach implemented on top of a Transformer Network.
\end{description}

\begin{table}[h]
    \centering
    \fontsize{7}{9}\selectfont
    \begin{tabular}{p{0.15\textwidth} | C{0.09\textwidth} | p{0.03\textwidth} p{0.03\textwidth} p{0.03\textwidth} }
    Baseline                & \scriptsize{Image$\vert$Label} & CIDEr & $\CovE$ & $\CovO$ \\ \hline
    \scriptsize{Labels-to-captions}          & N$\vert$Y & 62.08 & 21.01 & 6.19 \\
    \scriptsize{\cite{anderson2018bottom}}   & Y$\vert$N & 51.09 & 7.30 & 4.95 \\
    \scriptsize{\cite{sharma2018conceptual}} & Y$\vert$N & 62.35 & 10.52 & 6.74 \\
    \scriptsize{\cite{lu2018entity} w/ T}    & Y$\vert$Y & {\bf 69.46} & {\bf 36.80} & {\bf 6.93} \\
    \hline
    \end{tabular}
    \caption {
      Baseline model results, using either image or entity labels (2nd column).
      The informativeness metric $\CovE$ is low when additional input labels are not used, and high when they are.
    }
    \label{tab:quant-baselines}
\end{table}

Table~\ref{tab:quant-baselines} shows the performance of these baselines.
We observe that the image-only models perform poorly on $\CovE$ because they are unable to identify them from the image pixels alone.
On the other hand, the labels-only baseline and the proposal of \citet{lu2018entity} has high performance across all three metrics.

\begin{table}[h]
  \fontsize{8}{10}\selectfont
  \begin{center}
    \begin{tabular}{c|cc|ccc}
    Entity &
      $\WeE$ &
      $\WeO$  &
      CIDEr   &
      $\CovE$ &
      $\CovO$ \\ \hline
    Type  & \colorbox{yellow}{1.0}  & \colorbox{green}{1.0}  & 74.60 & \colorbox{yellow}{40.39} & \colorbox{green}{6.87} \\
    Type  & \colorbox{yellow}{1.5}  & 1.0  &  70.81 & \colorbox{yellow}{\bf{42.95}} & 7.04  \\
    Type  & 1.0  & \colorbox{green}{1.5}   &  73.82 & 40.03 & \colorbox{green}{8.38} \\
    Type  & 1.5  & 1.5                     &  71.11 & 41.94 & \bf{8.48} \\
    \hline
    Name  & \colorbox{yellow}{1.0}  & \colorbox{green}{1.0}  &  \bf{87.25} & \colorbox{yellow}{31.01} & \colorbox{green}{6.27} \\
    Name  & \colorbox{yellow}{1.5}  & 1.0  &  83.62 & \colorbox{yellow}{38.08} & 6.76 \\
    Name  & 1.0  & \colorbox{green}{1.5}   &  83.34 & 30.64 & \colorbox{green}{7.74} \\
    Name  & 1.5  & 1.5                     &  82.18 & \bf{38.17} & \bf{7.93} \\ \hline
    \end{tabular}
    \caption {
        Variants of our proposed approach using both image and entity labels as inputs.
        We present ablations on the coverage boost factors ($\WeE$ and $\WeO$) and entity-label modeling (type-only versus surface-form names).
        Informativeness of captions ($\CovE$ and $\CovO$) increases as coverage boost factors are increased (correlations highlighed in yellow and green).
    }
    \label{tab:quant}
    \end{center}
\end{table}

Table~\ref{tab:quant} shows the performance of our model.
Using both image and input labels improves performance on all metrics, compared to the baselines in Table~\ref{tab:quant-baselines}.
This indicates the effectiveness of our multi-encoder, multi-gated decoder architecture (\S~\ref{sec:multi-enc-transformer}) in generating captions that are both informative and fluent.
Moreover, boosting the weight for web entity labels ($\WeE$) and object labels ($\WeO$) improves informativeness for each of these types, see patterns highlighted in Table~\ref{tab:quant} for $\CovE$ and $\CovO$, respectively.\footnote{
Note that scaling weights at 2.0 or larger lead to repetition of input labels in the captions, resulting in fluency degradation without additional gains in informativeness.}
In terms of label modeling (type-only versus surface-form + type), the CIDEr score tends to be significantly higher when modeling surface-forms directly, while coverage metrics favor the type-only setting.
We attribute the former to the fact that the ground-truth captions are not very sensitive to label accuracy, 
and the latter to the fact that it is easier for the model to learn and generalize using the closed set of token types (approx. 4500 types).

We mention here that evaluating the performance of the image labeler models used (for object labels, entity recognition) is outside the scope of this work.
Their (possibly noisy) outputs are assumed given as input, and our evaluation measures the extent to which various image captioning models are capable of incorporating this information.

\subsection{Qualitative Results}

\begin{figure*}[ht!]
  \centering
  \scriptsize
  \renewcommand{\arraystretch}{1.1}
  \begin{tabular}{p{0.20\textwidth} p{0.32\textwidth} p{0.40\textwidth}}
     &
     \multicolumn{1}{l}{\includegraphics[width=0.20\linewidth]{imgs/sample.png}} &
     \multicolumn{1}{l}{\includegraphics[width=0.10\linewidth]{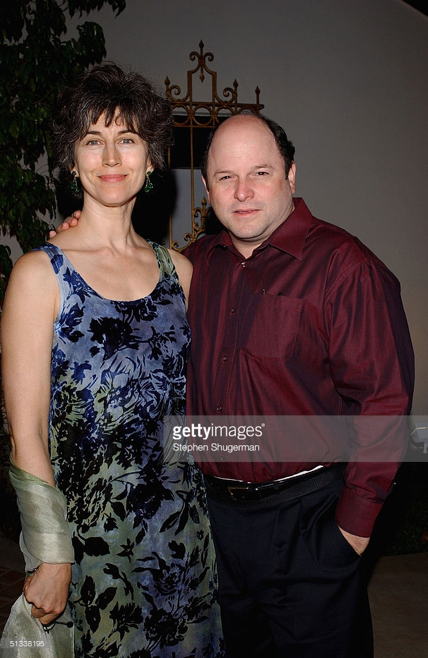}} \\
     \hline
     \textbf{\emph{Labels (as additional inputs)}} & & \\
     \hline
     \emph{Web Entity labels (WE) / \color{blue} Types} &
     \begin{tabular}[c]{@{}l@{}}eric clapton / \LabelType{ARTIST}\\ musician / \LabelType{PROFESSION}\\ crossroads guitar festival 2013 / \LabelType{ALBUM} \end{tabular}  &
     \begin{tabular}[c]{@{}l@{}}jason alexander / \LabelType{ACTOR}\\ daena e. title / \LabelType{PERSON}\\ geffen playhouse / \LabelType{THEATER}\end{tabular} \\
     \emph{Object labels} &
     guitarist, music artist, performance, stage, concert &
     lady, fashion, formal wear \\
     \hline
     \multicolumn{1}{l}{\textbf{\emph{Model Variants}}} & \multicolumn{1}{c}{\textbf{\emph{Output}}} & \multicolumn{1}{c}{\textbf{\emph{Output}}} \\
     \hline
     \emph{Image only} &
     person performs on stage &
     people arrive at the premiere \\
     \emph{Image + WE Types} &
     \LabelType{ARTIST} performs on stage &
     \LabelType{ACTOR} and \LabelType{PERSON} attend the opening night \\
     \emph{(Above) + Postprocessing} &
     eric clapton performs on stage &
     jason alexander and daena e. title attend the opening night \\
     \emph{Image + WE Types + 1.5 Boost} &
     \LabelType{ARTIST} performs live during a concert &
     \LabelType{ACTOR} and \LabelType{PERSON} attend the premiere at \LabelType{THEATER} \\
     \emph{(Above) + Postprocessing} &
     eric clapton performs live during a concert &
     \textbf{jason alexander and daena e. title attend the premiere at geffen playhouse } \\
     \emph{Image + WE} &
     eric clapton performs live during a concert &
     actor jason alexander and wife  daena title arrive at the geffen playhouse premiere \\
     \emph{Image + WE + 1.5 Boost} &
     \textbf{eric clapton performs on stage during the crossroads guitar festival} &
     jason alexander and \textcolor{red}{daena e. title daena playtitle} attend the premiere at geffen playhouse \\
     \hline
  \end{tabular}
  \caption{Sample outputs for various model configurations for two images and their additional label inputs. In both cases, we notice that boosting the coverage at inference time leads to more informative captions without a loss in fluency. 
  Note: Object labels are provided as inputs to the model in all cases except the baseline {\it Image only}.}
  \label{fig:sample-outputs}
\end{figure*}

To get a better intuition on the behavior of our models, we compare output captions of different model variants using two sample images (Fig.~\ref{fig:sample-outputs}).
The baseline model without any input labels tends to produce generic-sounding captions (i.e., refer to the people in the image simply as `person').
When we supply web entity types as inputs, the outputs become more informative as evident from the use of output types,  e.g. \LabelType{ARTIST}, which in turn is postprocessed to match the web entity label ``eric clapton''.
Furthermore, increasing the Coverage Boost Factor to 1.5 (i.e., both $\WeE$ and $\WeO$ set to 1.5) results in more informative captions that add previously-missing aspects such as ``concert'' and ``geffen playhouse''.

Similar trends are seen with direct modeling of the fine-grained labels for web entities.
While successfully adding additional information under the Coverage Boost 1.5 condition for the first image (``the crossroads guitar festival''), we observe an error pattern this model exhibits, namely, the presence of ill-formed named entities in the output (``daena e. title daena playtitle'', indicated in red in Fig.~\ref{fig:sample-outputs}).
The human evaluation results show that this model configuration performs worse compared to the one using entity types only, which are both easier to learn by the model and guaranteed to preserve the full name of the entity as copied during postprocessing from the input labels.

Please see Fig.~\ref{fig:more-qualitative} for more test images and output captions.
Our model generates captions that fluently incorporate fine-grained entity mentions that are provided as input labels, e.g. ``troms\o'' (city in northern Norway), ``basmati'' (type of rice), ``aruba'' (island in the Caribbean Sea) and ``kia ceed'' (model of car).
In the cases where such specific details are not available as inputs, the model uses a generic term to describe the scene, e.g. ``\textbf{musician} playing the saxophone on stage''.

\begin{figure}[h]
\begin{center}
 \includegraphics[width=0.99\linewidth]{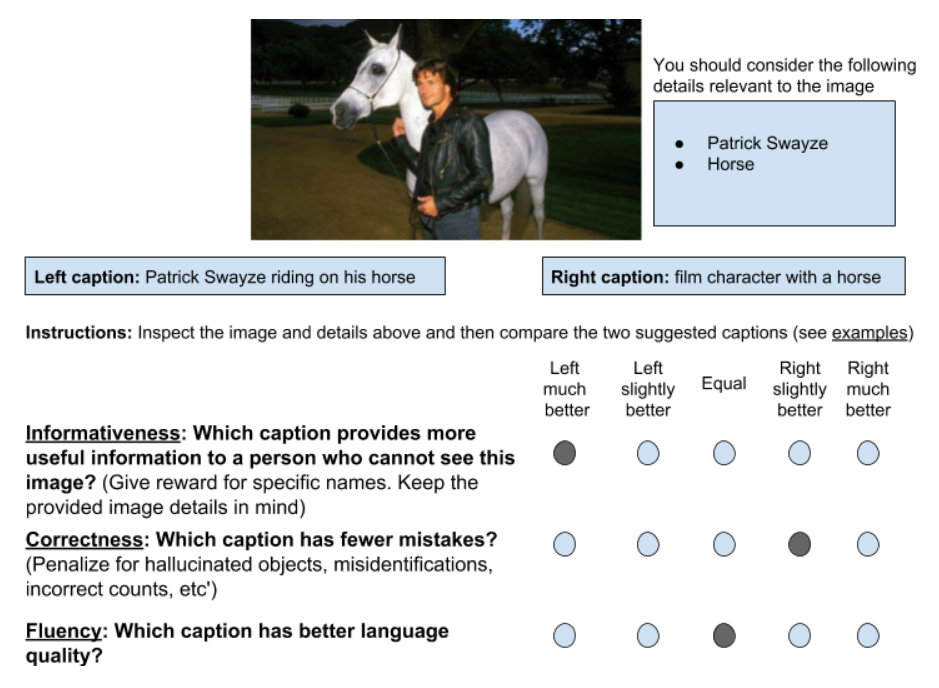}
\end{center}
 \caption{Interface for the human evaluation.}
\label{fig:human_eval_plugin}
\end{figure}

\subsection{Human Evaluation}
\label{sec:human_eval}

We conducted a human evaluation study to determine whether the gains on label coverage scores (Table~\ref{tab:quant}) correlate with accuracy as judged by humans.
Hence we used the models with high coverages of object labels and web entities in Table~\ref{tab:quant}, which correspond to 1.5 Coverage Boost.
Each of our proposed models was independently compared in a side-by-side evaluation (randomized order) against the same baseline model on three different dimensions: Informativeness, Correctness and Fluency.
We used the model by \newcite{sharma2018conceptual} as the baseline model because the goal was to measure gains obtained from using the input labels,
and this baseline performed best (in terms of CIDEr score) amongst baselines not using input labels (Table~\ref{tab:quant-baselines}).

The evaluation setup and a description of each of the evaluation dimensions is given in Fig.~\ref{fig:human_eval_plugin}.
Note that the web entities that are fed to the model were also provided to the raters as reference (to help with fine-grained identification).
In this example, the left caption is judged higher on the Informativeness scale because it correctly identifies the person in the image,
but it is rated lower on the Correctness dimension due to the incorrect action (``riding'');
both captions are judged as equally fluent.

In each evaluation, three raters evaluate a 2K random sample batch from the test set.
The human ratings were mapped to the corresponding scores using the following scheme:
\begin{table}[H]
    \centering
    \scriptsize
    \begin{tabular}{l|c}
        The baseline caption is much better & -1.0  \\
        The baseline caption is slightly better & -0.5 \\
        The two captions seem equal & 0 \\
        Our model's caption is slightly better & +0.5 \\
        Our model's caption is much better & +1.0
    \end{tabular}
\end{table}
Table~\ref{tab:human_eval} reports the improvements in human evaluations using our setup against the baseline captions.
We observe that the gains in coverage scores in Table~\ref{tab:quant} are now reflected in the human judgements,
with the best model (using labels, type-only) judged as $24.33\%$ more informative and $7.79\%$ more correct, with virtually no loss in fluency.
Furthermore, these results validate the claim from \newcite{sharma2018conceptual} that generating captions containing fine-grained entities from image pixels only (without additional fine-grained labels)
leads to inferior performance in both informativeness (-$7.91\%$) and correctness (-$7.33\%$) (second row in Table~\ref{tab:human_eval}).

\begin{table}[h]
\begin{center}
\fontsize{8}{10}\selectfont
\begin{tabular}{llll|rrr}
\hline
L & T & {\tiny $\WeE$} & {\tiny $\WeO$} & Info' & Correct'   & Fluency \\ \hline
\multicolumn{4}{l|}{\newcite{sharma2018conceptual}} & 0.00\% & 0.00\% & 0.00\% \\
No & - & -  & - & -7.91\%  & -7.33\% & 0.11\% \\
\multicolumn{4}{l|}{\newcite{lu2018entity} w/ T}  & 7.45\% & 2.60\%  & {\bf 2.47}\% \\
Yes & No & 1.5 & 1.5 & {16.18\%} & {7.94\%} & {-0.06\%} \\
{\bf Yes} & {\bf Yes} & {\bf 1.5} & {\bf 1.5} & {\bf 24.33\%}  & {\bf 7.79\%} & -0.87\% \\
\hline
\end{tabular}
\caption{
  Side-by-side human evaluation results (first entry is the system used in all the comparisons).
  First column (L) indicates if entity-labels are used.
  Second column (T) indicates if entity type is used instead of the surface-form.}
\label{tab:human_eval}
\end{center}
\end{table}

\begin{figure*}[ht!]
  \centering
  \scriptsize
  \begin{tabular}{p{0.08\textwidth} p{0.25\textwidth} p{0.25\textwidth} p{0.25\textwidth}}
    &
    \hfill\includegraphics[height=0.75\linewidth,width=\linewidth]{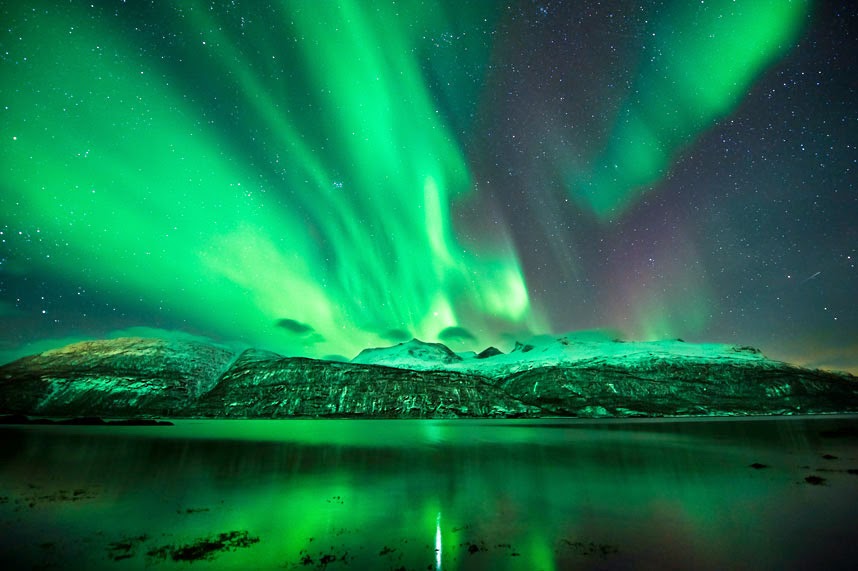}\hspace*{\fill} &
    \hfill\includegraphics[height=0.75\linewidth,width=\linewidth]{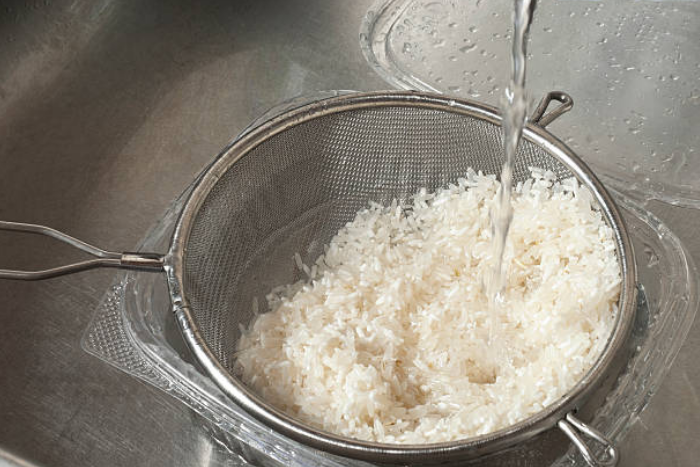}\hspace*{\fill} &
    \hfill\includegraphics[height=0.75\linewidth,width=\linewidth]{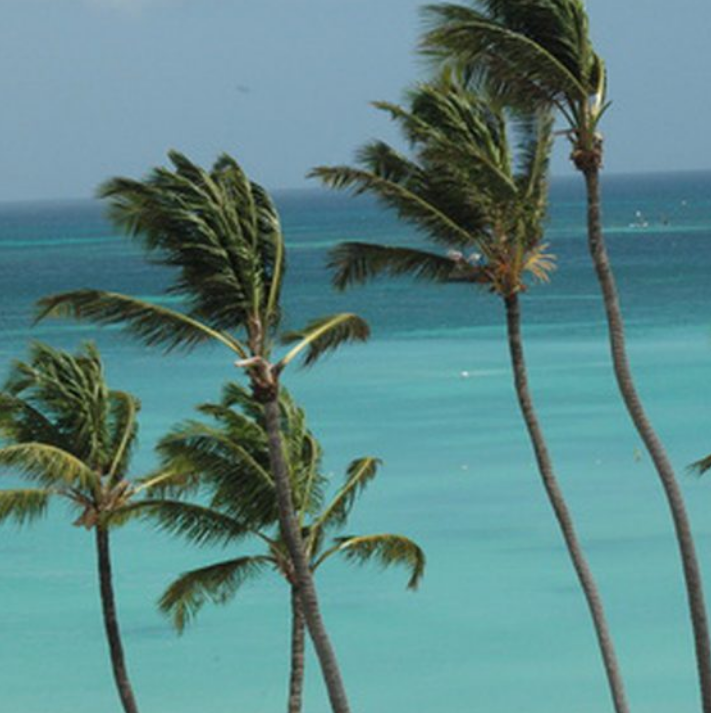}\hspace*{\fill} \\
    \hfill\textbf{Our Model} &
    aurora borealis over tromsø &
    basmati cooked rice in a pan &
    palm trees on the beach in aruba \\
    \hfill\textbf{Baseline} &
    the northern lights dance in the sky &
    white rice in a bowl &
    palm trees on the beach \\
    & & & \\
    &
    \hfill\includegraphics[height=0.75\linewidth,width=\linewidth]{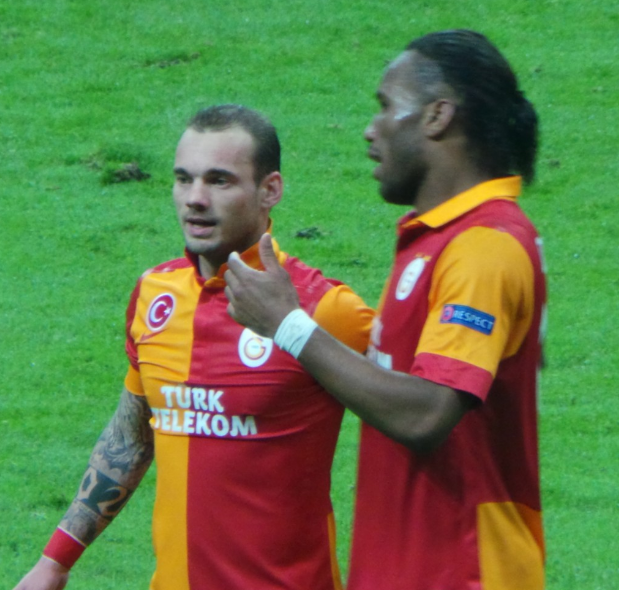}\hspace*{\fill} &
    \hfill\includegraphics[height=0.75\linewidth,width=\linewidth]{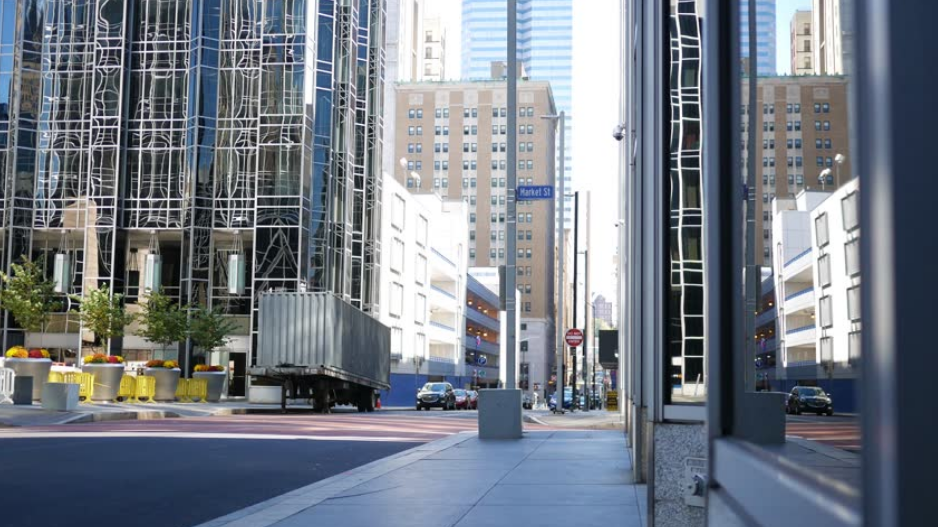}\hspace*{\fill} &
    \hfill\includegraphics[height=0.75\linewidth,width=\linewidth]{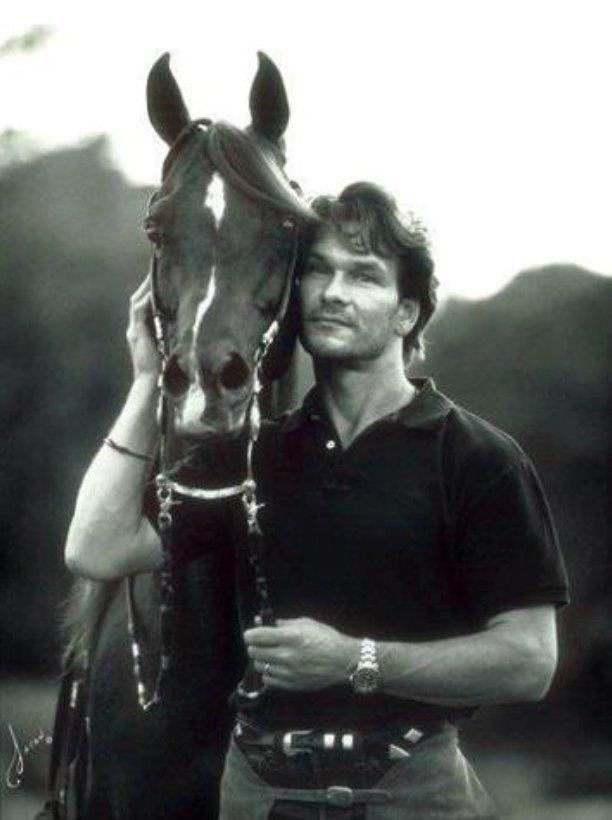}\hspace*{\fill} \\
    \hfill\textbf{Our Model} &
    didier drogba of phoenix rising fc celebrates with teammates . &
    chicago , traffic in the downtown of chicago &
    patrick swayze riding a horse in dirty dancing \\
    \hfill\textbf{Baseline} &
    person celebrates scoring his side 's first goal of the game &
    view from the southwest - full - height view &
    film character with a horse \\
    & & & \\
    &
    \hfill\includegraphics[height=0.75\linewidth,width=\linewidth]{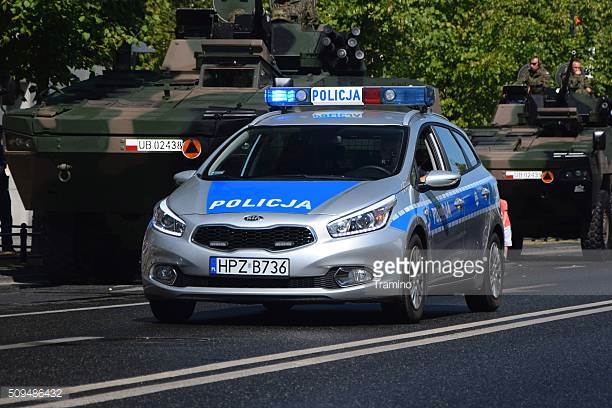}\hspace*{\fill} &
    \hfill\includegraphics[height=0.75\linewidth,width=\linewidth]{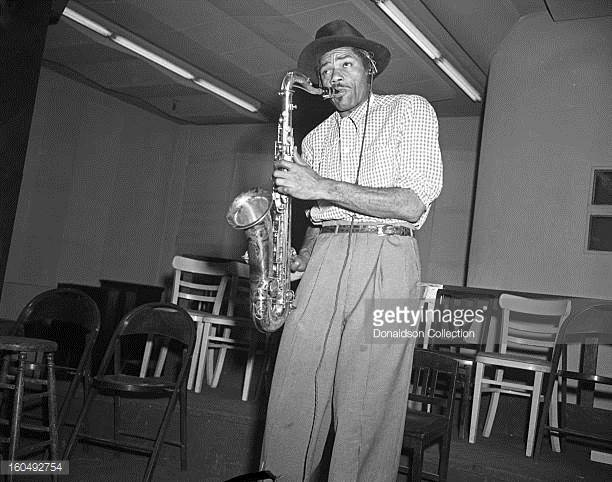}\hspace*{\fill} &
    \hfill\includegraphics[height=0.75\linewidth,width=\linewidth]{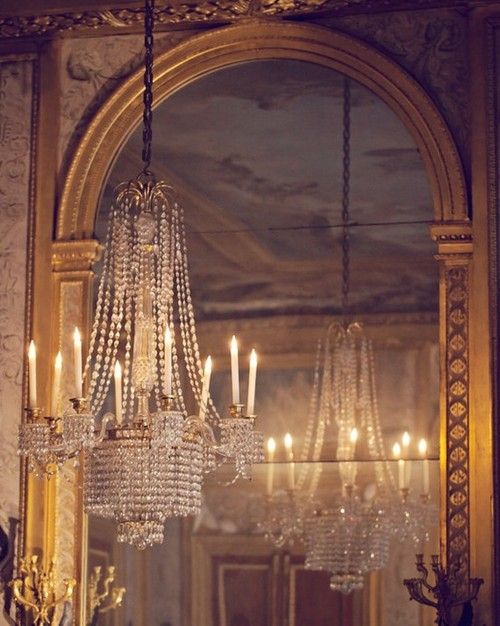}\hspace*{\fill} \\
    \hfill\textbf{Our Model} &
    kia ceed police car on the street &
    musician playing the saxophone on stage &
    candelabra chandelier in the lobby \\
    \hfill\textbf{Baseline} &
    police car on the street &
    jazz artist poses for a portrait &
    beautiful lighting in the hallway \\
    & & & \\
  \end{tabular}
  \caption{Qualitative results comparing baseline captions \cite{sharma2018conceptual} with our model that use web entity types and Coverage Boost Factor of 1.5 (i.e., both $\WeE$ and $\WeO$ set to 1.5).}
  \label{fig:more-qualitative}
\end{figure*}

\section{Conclusion}
We present an image captioning model that combines image features with fine-grained entities and object labels, and learns to produce fluent and informative image captions.
Additionally, our model learns to estimate entity and object label coverage, which can be used at inference time to further boost the generated caption's informativeness without hurting its fluency.

Our human evaluations validate that training a model against ground-truth captions containing fine-grained labels (but without the additional help for fine-grained label identification), leads to models that produce captions of inferior quality.
The results indicate that the best configuration is one in which fine-grained labels are precomputed by upstream models, and handled by the captioning model as types, with additional significant benefits gained by boosting the coverage of the fine-grained labels via a coverage control mechanism.


\bibliography{informative_captions}
\bibliographystyle{acl_natbib}

\end{document}